\documentclass[runningheads]{llncs}
\usepackage[T1]{fontenc}
\usepackage{graphicx}
\usepackage{caption}
\usepackage{subcaption, wrapfig}
\usepackage{multirow}
\usepackage{float}
\usepackage{placeins}
\usepackage{tabularx,booktabs}
\usepackage{array}
\usepackage{color}
\usepackage{soul}
\usepackage[dvipsnames]{xcolor}
\usepackage[framemethod=tikz]{mdframed}
\newcolumntype{P}[1]{>{\centering\arraybackslash}p{#1}}
\usepackage[justification=centering]{caption}
\usepackage[dvipsnames]{xcolor} 

\begin{document}

\raggedbottom

\title{Cross-Attention Based Influence Model for Manual and Nonmanual Sign Language Analysis}
\titlerunning{Cross-Attention Based Influence Model}
%
 \author{Lipisha Chaudhary\inst{1\thanks{Corresponding author: lipishan@buffalo.edu}}\orcidID{0000-0002-8863-1338} \and \\
Fei Xu\inst{1} \orcidID{0000-0002-9353-9528} \and \\
Ifeoma Nwogu\inst{1}\orcidID{0000-0003-1414-6433}}
\authorrunning{Chaudhary et al.}
%
\institute{Department of Computer Science and Engineering, University at Buffalo, \\ Buffalo, New York, USA \\
\email{\{lipishan, fxu3, inwogu\}@buffalo.edu}}
\maketitle              
\begin{abstract}
Both manual (relating to the use of hands) and non-manual markers (NMM), such as facial expressions or mouthing cues, are important for providing the complete meaning of phrases in American Sign Language (ASL). Efforts have been made in advancing sign language to spoken/written language understanding, but most of these have primarily focused on manual features only. In this work, using advanced neural machine translation methods, we examine and report on the extent to which facial expressions contribute to understanding sign language phrases. We present a sign language translation architecture consisting of two-stream encoders, with one encoder handling the face and the other handling the upper body (with hands). We propose a new parallel cross-attention decoding mechanism that is useful for quantifying the influence of each input modality on the output. The two streams from the encoder are directed simultaneously to different attention stacks in the decoder. Examining the properties of the parallel cross-attention weights allows us to analyze the importance of facial markers compared to body and hand features during a translating task.

\end{abstract}
%
%
%
\FloatBarrier\section{Introduction}
\label{intro}

Sign Language is the primary means of communication used by Deaf and Hard of Hearing (DHH) individuals, uses multimodal cues to fully express the intended meaning. In the literature, it has been frequently noted that the linguistics of many sign languages have yet to be as well analyzed and studied when compared to their spoken counterparts \cite{c843170b-f7b9-3a6d-86cf-0f127d816a9f}. This makes exploring automated sign language understanding an important research field where emphasis can be put on analyzing how other visual cues support the sign gestures \cite{Bragg2019SignLR}.

Similar to spoken language, sign language has its own syntax, grammar, and semantics. Spoken language uses speech as one of its main vehicles of transmission. On the contrary, sign language combines visual gestures such as hand shapes and movements, body movement, mouthing cues, and facial expressions. These components are divided into two major categories: a) manual markers, which include parameters like hand shapes, palm orientation, hand/arm movement, and hand location changes, and b) non-manual markers, such as facial expressions and mouthing cues.

Current works in sign language translation or generation mainly use only the manual markers, oftentimes ignoring the non-manual aspects of the sign or, at best, including them implicitly \cite{8578910}. While this may be sufficient to accomplish simple comprehension tasks, they lack the ability to capture the rich expressive power\footnote{The expressive power of a language refers to the variety and quantity of concepts that can be represented and communicated in that language.} of the sign language. 

In this work, we are interested in analyzing the importance of using manual and non-manual markers in Sign Language Translation (SLT). 
While there have been different transformer-based models in the general pattern recognition/ machine learning community that model multimodal signals by fusing them and then decoding the fused embedding, such models are unable to disentangle the individual contributions of each input modality to the output result. Fusion happens, but the extent of the contributions becomes opaque due to the construction of such models.

Hence, in this work, we aim to take advantage of how the cross-attention or (encoder-decoder attention) mechanism in transformers highlights the relationship between the input feature tokens and the plausible predicted output tokens. Using such attention we can easily understand which input feature attends to which output token the most at time $t$. 
Our proposed model encodes each input channel separately and then fuses them via \emph{cross-attention in the decoder}. This ensures that each feature embedding is used to predict the final output tokens, thus providing insight into the extent of the individual feature contributions. To the best of our knowledge, there are no such multimodal transformer architectures that can be used to explicitly measure the influence of each encoder input on the resulting decoder output.

We evaluate our proposed method by estimating the influence of manual and nonmanual features during Sign Language translation, tested on the RWTH-PHOENIX-Weather2014T (PHOENIX14T) dataset \cite{8578910} and the real-life American Sign Language dataset (ASLing) \cite{ananthanarayana2021dynamic}.
More details on the datasets are given in Section \ref{dataset}. We benchmark our method against other translation methods that use PHEONIX14T, thus demonstrating the viability of the proposed method. We also use the real-life ASLing dataset to provide a more qualitative assessment.

In summary, we propose a novel parallel cross-attention decoder transformer-based approach to analyze the contributions of each component, hand-based sign gestures (manual) and facial expressions (non-manual markers), in the Sign Language translation (SLT) task. While significant efforts have been made to improve the performance numbers in SLT tasks, in this work, our focus remains on conducting a comparative analysis of two major sign language components and how they influence the downstream task of translation.

\let\labelitemi\labelitemii

\noindent To this end, the contributions of this work can be summarized as follows:
\begin{itemize}
    \item We design a general, all-purpose, multi-channel cross-attention (encoder-decoder attention) fusion model useful in multimodal pattern analysis for understanding the influence that each input modality contributes to the resulting end task.

    \item Using the proposed cross-attention fusion technique, we analyze the role of facial expression in the Sign Language translation task.

    \item Employing two well-established datasets in the field, we present quantitative and qualitative evaluation approaches for a continuous sign translation task. These rely on the attention weights created by the parallel cross-attention model during inference.

\end{itemize}

\FloatBarrier\section{Related Works}

\paragraph{\textbf{Multi-Modal Sign Language Translation}}
Sign Language Translation (SLT) focuses on interpreting the signals conveyed in signing videos to spoken/written text. Early works \cite{8578910} introduced an RNN-based pipeline to solve the SLT problem. Many works \cite{camgoz2020sign,li2020tspnet,NIPS2017_3f5ee243,8578910} began using Transformer-based architectures to further improve the performance. Most of these works have used skeleton-based features focusing on the upper body and hands. Camgoz et al. \cite{10.1007/978-3-030-66823-5_18} introduced a framework that allowed for separate input channels to process individual sign language components. Other works made similar advances by also incorporating multiple modalities, though not necessarily multiple components of Sign Language \cite{chen2022twostream} - their two modality streams were the (i) raw RGB frames and (ii) corresponding body keypoint sequence extracted from the frames. Chaudhary et al. \cite{9999492} built an end-to-end 2-way pipeline to use the sign-translated phrases to generate the original signs.

\paragraph{\textbf{Manual and Non-Manual Markers}}

There is a common misconception that Sign Language is only communicated using hand gestures \cite{20143163}, but in reality, it is much more complex than this. Similar to phonemes in sound, Sign Language phonology underscores the structure of each sign and the way they are organized. 
Each sign can be broken down into smaller parts made up of the handshape, hand location, hand/arm movement, palm orientation, and the corresponding nonmanual cues \cite{af764da8-3e31-3e1d-9d2c-d0f4f601d03d}.
As outlined in Section \ref{intro}, this work focuses on understanding the manual signals (especially the upper torso with handshapes and their movements) and non-manual signals, specifically facial expressions as related to sign understanding. \medskip

Silva et al. \cite{9320274} introduced a FACS-based facial expression database for Brazilian Sign Language. Mukushev et al. \cite{mukushevevaluation} analyzed similar signs to find if non-manual components can differentiate them distinctly. Zheng et al. \cite{ZHENG2021462} attempted to improve the performance of SLT by adding facial expressions as input. Koller et al. \cite{7406418} modeled mouthing shapes in reference to the sign language recognition task.

\section{Method}
We introduce a fusion model that is useful for understanding the weightage of each input feature representation during the training process when the model is presented with more than one type of input. The proposed model consists of two encoders and a decoder. The architecture is trained for a Sign Language Translation (SLT) task, which is considered a sequence-to-sequence learning problem \cite{8578910}.

\begin{figure}
    \begin{center}
    \includegraphics[width=\textwidth,height=17cm, keepaspectratio]{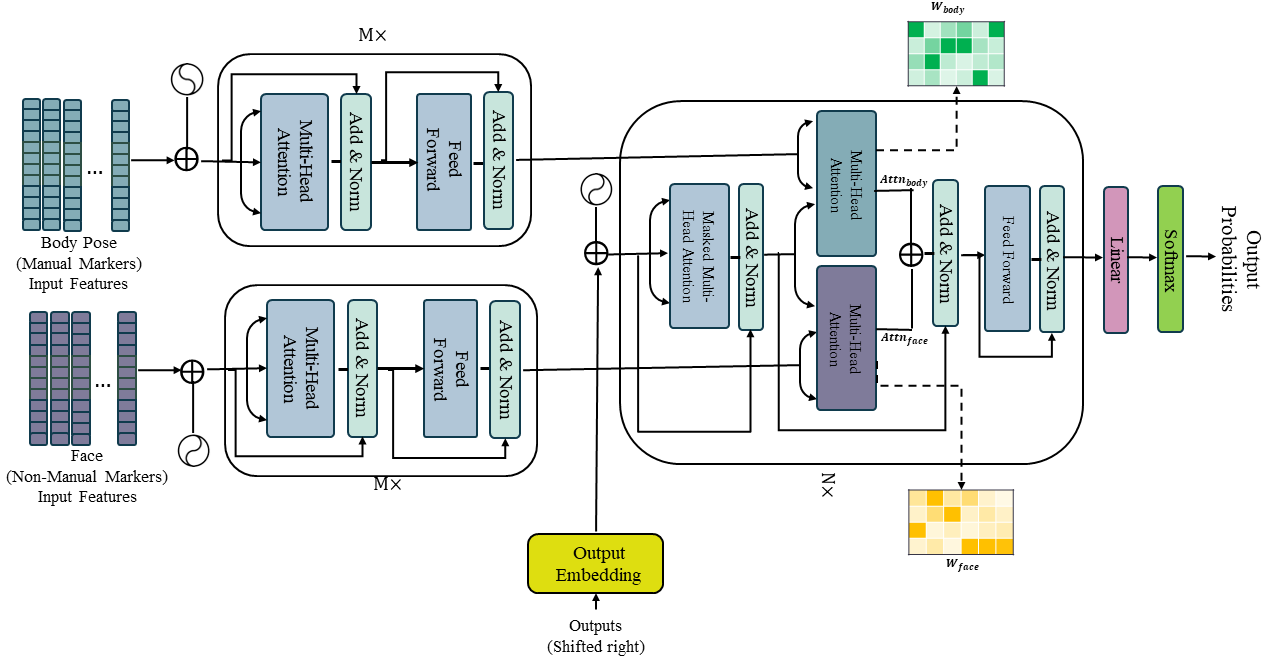}
    \caption{Overview of the proposed architecture showing the two-stream dual encoder and the dual-cross-attention based decoder.}\label{fig:overview}
    \end{center}
\end{figure}

Figure \ref{fig:overview} shows the overview of our proposed architecture, which consists of a) two separate encoders, each for a single feature input, and b) a dual-cross-attention-based decoder for generating spoken language output texts.

\subsection{Parallel Cross-Attention Decoder Transformer}
\label{model}
The proposed architecture follows the framework of a standard encoder-decoder Transformer but consists of two detached encoders and an added encoder-decoder attention layer. Each encoder is responsible for learning one input feature representation: (i) hands and body pose (manual markers) and (ii) face features (non-manual markers). The model is trained to accomplish the task of SLT by learning the attention weights of individual features using a dual-cross-attention module as the intermediary task. The intuition is that the problem at hand involves understanding how much each of the two separate inputs contributes to the single final task. A similar architecture \cite{Le2020DualdecoderTF} used a dual-decoder-based transformer for the joint simultaneous tasks of automatic speech recognition (ASR) and speech translation. Note that this significantly differs from our proposed work where we have dual encoders interacting within the cross-attention block.

By simultaneously having two separate attention blocks being responsible for each of the input features, the properties of each attention block can be evaluated to determine how much influence that encoder (hence that modality) has on the final output decoder task.  

\subsubsection{Two-stream encoding pipeline}

We make use of two separate encoders to independently learn the representation of each of the input features. Each encoder consists of an input embedding layer followed by a positional embedding and several self-attention and feed-forward network (FFN) layers whose inputs are normalized. Both the encoders follow the same layer configuration as a standard Transformer encoder \cite{NIPS2017_3f5ee243}. 

The first encoder takes the segments of stacked 3D body landmarks \cite{lin2023one} $X_b = ({x_1}^b, {x_2}^b, {x_3}^b, ...., {x_n}^b)$  (further explained in Section \ref{dataset}) and the second encoder takes segments of stacked 3D face landmarks \cite{lin2023one} $X_f = ({x_1}^f, {x_2}^f, {x_3}^f, ...., {x_n}^f)$ as input sequence. Similar to the standard encoder layers the input sequence is modeled with self-attention and mapped into contextual representation: $z_b = ({z_1}^b, ..., {z_n}^b)$ for the manual markers which are the body and hand joints\footnote{Here landmarks and joints can be used interchangeably} and $z_f = ({z_1}^f, ..., {z_n}^f)$ for the face landmarks which are the non-manual markers. Each encoder stream is composed of $M$ number of layers.

\subsubsection{Decoding pipeline}

The parallel cross-attention decoder consists of a single decoder with an additional multi-head attention layer. The additional multi-head attention layer paired with the existing multi-head attention layer together are called the \textbf{parallel cross-attention layers}. Each parallel attention layer follows a standard encoder-decoder attention layer (also called multi-head attention). An attention layer is defined by a function that maps a query and a pair of key-value to an output vector. The attention function receives the inputs as a key $K$, a value $V$, and a query $Q$. The decoder usually performs two types of attention functions: a) a self-attention is performed on the shifted output sequence, where the inputs $K$, $V$, and $Q$ are the same, and b) a cross-attention, also known as encoder-decoder attention, which maps the context representations with the output sequence that are received from the self-attention layer. 

In order to understand the importance of any feature against the output sequence, cross-attention values are considered. Taking inspiration from parallel combination strategy \cite{Libovick2018InputCS}, we propose a model that will learn the context representations from both features simultaneously, along with the weightage the model puts on each feature when deciding on the final output sequence. To formulate the problem, we consider the context representations from both encoders as separate inputs to their respective attention function. 

The attention function is given as:
\begin{equation} \label{eq:1}
    Attn(Q, K, V) = softmax \left(\frac{QK^{T}}{\sqrt{d_{k}}}\right)V
\end{equation}
In the proposed parallel cross-attention decoder, each attention function is responsible for attending to the output context representation from each encoder.  We let $Attn_{body}$ be the attention layer that takes the output context representation $z_b$ of the body encoder as the key-value pair input. Similarly, $Attn_{face}$ is the attention layer that attends to the output context representation $z_f$ from the face encoder as the key-value pair. In both cases, the query value is received by the normalized masked self-attention output from the previous shifted token of the sequence.

To formulate the concept of parallel cross-attention decoder, we denote the inputs as follows:

$Q_{shifted} $ is the query value received from the shifted output sequence

$K_{(z_f)}, V_{(z_f)}$ is the key-value pair received from the face encoder stream

$K_{(z_b)}, V_{(z_b)}$ is the key-value pair received from the body encoder stream

\noindent Thus, each of the attention functions in the parallel cross-attention decoder is:

\begin{equation} \label{eq:2}
    Attn_{body} = softmax \left(\frac{Q_{shifted}K_{(z_b)}^{T}}{\sqrt{d_{k}}}\right)V{(z_b)}
\end{equation}

\begin{equation} \label{eq:3}
    Attn_{face} = softmax \left(\frac{Q_{shifted}K_{(z_f)}^{T}}{\sqrt{d_{k}}}\right)V{(z_f)}
\end{equation}

The final output from the parallel cross-attention is received by merging both the attention outputs from \ref{eq:2} and \ref{eq:3}. We denote the final merged output as $Attention_{final}$. The concatenation operator was used to merge the outputs to get to the final output:

\begin{equation} 
    Attn_{final} = linear([Attn_{body}; Attn_{face}])
\end{equation}

The combined attention output is normalized and then passed onto the feed-forward network to predict the next token auto-regressively. We simultaneously look at the attention weights from both the attention functions for the body and face. For this, we focus on the correlation computed between the query ($Q_{shifted}$) and each key value - $K_{body}$ \& $K_{face}$. Specifically, we observe the outputs of the Scaled Dot-Product Attention \cite{NIPS2017_3f5ee243} after the softmax layer. We formulate the attention weights as follows:

\begin{equation} \label{eq:5}
    \small{w_{body} = softmax \left(\frac{Q_{shifted}K_{(z_b)}^{T}}{\sqrt{d_{k}}}\right)}
\end{equation}

\begin{equation} \label{eq:6}
    \small{w_{face} = softmax \left(\frac{Q_{shifted}K_{(z_f)}^{T}}{\sqrt{d_{k}}}\right)}
\end{equation}

We use the weights from Equations \ref{eq:5} and \ref{eq:6} to analyze the role of manual and non-manual markers in understanding sign language and present our findings in Section \ref{exp}.

\FloatBarrier\section{Implementation and Evaluation Details}
\FloatBarrier\subsection{Datasets}
\label{dataset}
First, we use the benchmark dataset, PHOENIX14\textbf{T} \cite{8578910} to evaluate and establish credibility for our proposed translation model. We train our proposed parallel cross-attention decoder transformer with 7096 training, 519 validation, and 642 test samples. The samples were collected from the weather forecast airings and performed by nine different signers in German Sign Language (GSL)\footnote{German Sign Language is also known as Deutsche Gebärdensprache (DGS).}. Along with the samples, the dataset contains their corresponding German translations and gloss annotations.

Next, we evaluate the proposed model on the more real-life, unconstrained American Sign Language dataset, ASLing \cite{ananthanarayana2021dynamic}. This dataset consists of 1027 training samples and 257 testing samples. The samples were performed by 7 native signers and collected at 10 frames per second. 

\begin{figure}[H]
    \centering
    \includegraphics[width=0.65\textwidth]{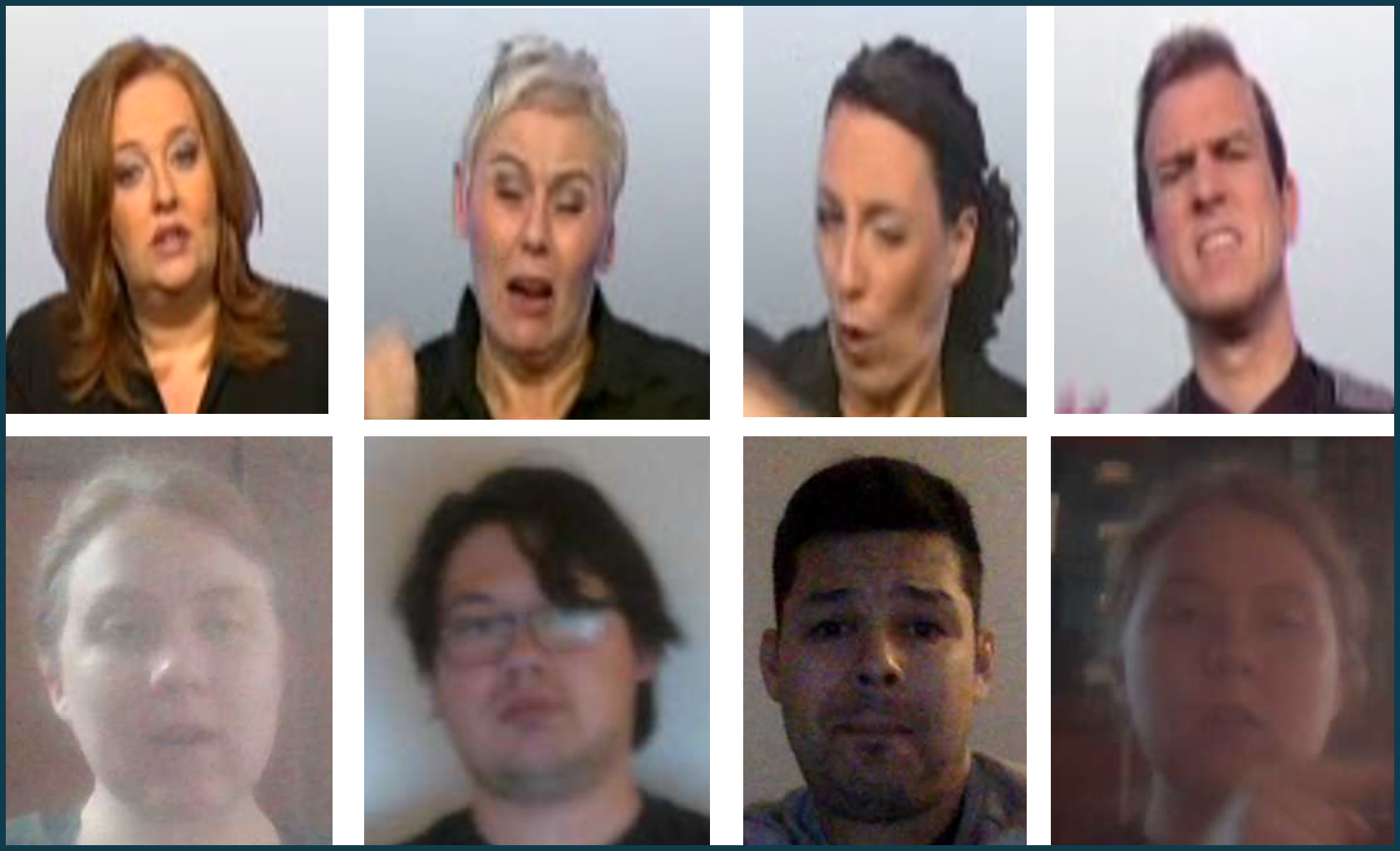}
    \caption{Face crop samples from the datasets. Top: from the Phoenix2014 dataset; \\
    Bottom: from the ASLing dataset}
    \label{fig:face-crops}
\end{figure}

\subsection{2-stage Input Feature Extraction Process}
\begin{wrapfigure}{5}{0.5\textwidth} 
\vspace{-20pt}
\centering
\begin{tabular}{c c}
    \includegraphics[width=0.5\linewidth]{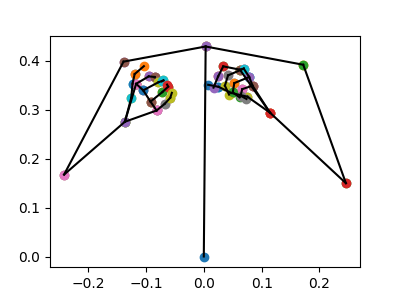} &
    \includegraphics[width=0.5\linewidth]{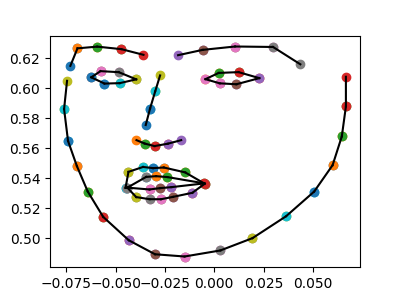}
\end{tabular}
\caption{ The $(x, y)$ joint plots for body (left) and face (right). Note that we use 3D points $(x, y, z)$, in our analysis.}\label{fig:landmarks}
\vspace{-18pt}
\end{wrapfigure}

\paragraph{\textbf{3D keypoints}} To accurately track the motion of each of the markers (body and face), we make use of 3D joint features $j = (x, y, z)$. We use \mbox{\cite{lin2023one}} to extract both body and face 3D joints for the input sequence $X_M$. To independently understand the weightage of each of the modalities, we separate the body and face joints for individual video input sequences. Given the 3D graphical representations of the input features, we employ a Spatial-Temporal Graphical Convolution Network (STGCN) \mbox{\cite{yu2018spatio}}. Indices are selected based on points depicted in Figure \ref{fig:landmarks}.
%

For the manual markers, we consider only the upper torso with hands. The body keypoints follow \cite{SMPL-X:2019} format; we eliminate the foot, ankle, and knee keypoints to get a total of $48$ body keypoints including hands. We consider all $72$ face keypoints for the non-manual markers to create the skeleton. There is a total of $120$ selected keypoints which are used as the input to the embedding extractor. To build a connection between the body and face keypoints, we build a custom tree structure which is used in the processing of a Spatio-Temporal Graphical Convolution Network (STGCN). 

\begin{figure}[H]
    \centering
    \includegraphics[width=\textwidth]{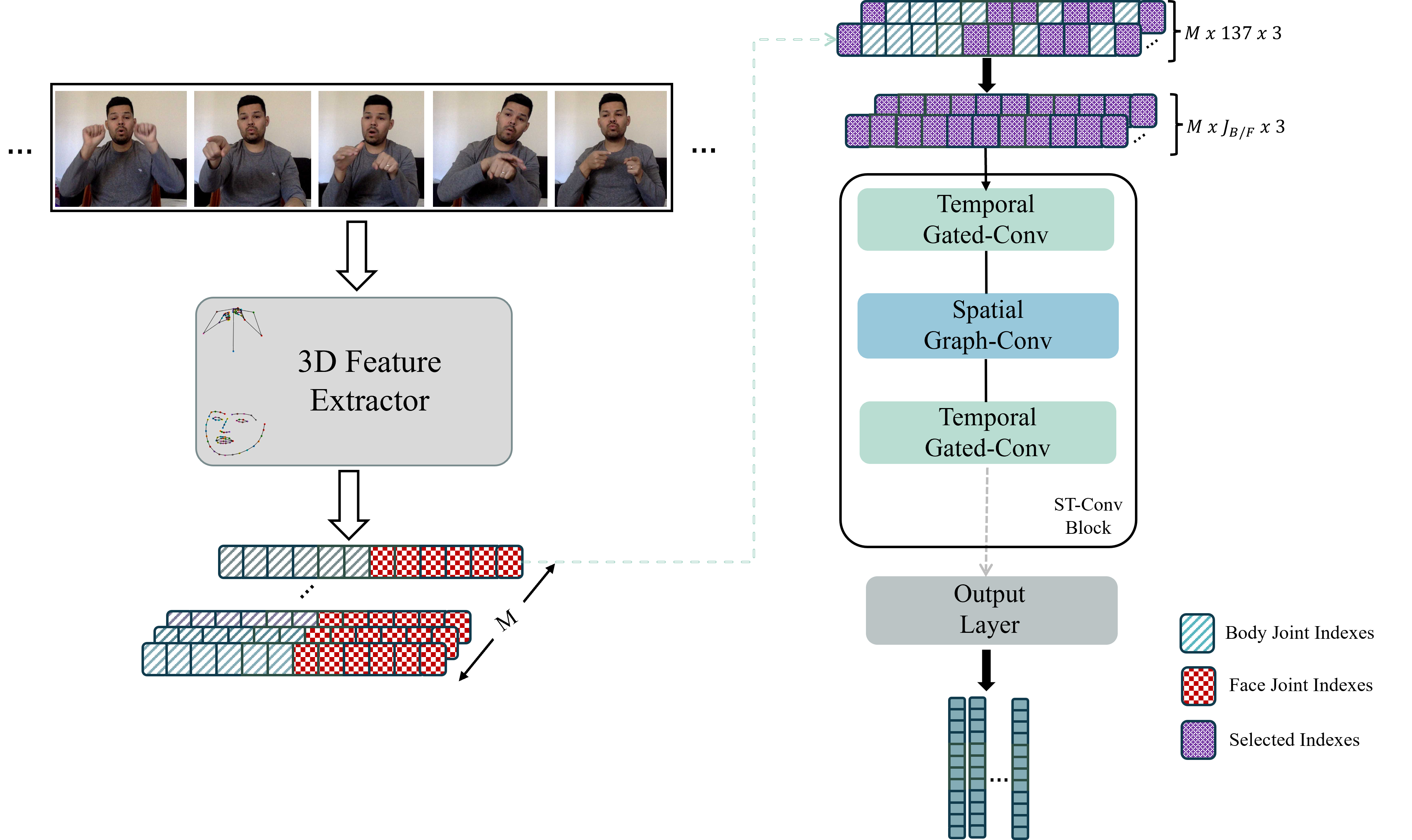}
    \caption{Overview of the 2-stage input feature extraction process.
    }
    \label{fig:ovinp}   
\end{figure}

We pre-train the STGCN on a word-level Sign Language recognition dataset (WLASL) \cite{li2020word} in order to learn general sign language motion representations.  The resultant vector per temporal window is of length $1024$. Figure \ref{fig:ovinp} shows the process of feature extraction. The right side of the process is pretrained twice; once on manual input features and the second on non-manual input features, to get $1024$ feature embeddings per modality.

\paragraph{\textbf{Rotation Matrix}} Another feature we consider with the human skeleton structure is the 3D rotation matrix, a special orthogonal $3\times 3$ matrix ($SO(3)$ ), that represents the rotation of one joint in Euclidean space. A chain of rotation matrices from the origin joint (pelvis) on the skeleton is used to represent the human pose for each video frame.

Recent works \cite{levinson2020analysis} \cite{brunton2021modern} have shown the effectiveness of different rotation representations of $SO(3)$ for gradient-based optimization in neural network learning. Specifically, a rotation representation that contains less than five dimensions suffers from discontinuities in the real Euclidean space, which leads the gradient of the loss function to blow up \cite{geist2024learning}. The $3\times 3$ rotation matrix we considered in this work contains 9 parameters, which bakes in the redundancies needed to deal with the discontinuous issue as compared to other rotation representations. 

To make a fair comparison with 3D keypoint features, we follow the same protocol to consider only the upper body and hand joints. We also pre-train the STGCN with the WLASL dataset to learn the best motion representations from the rotation matrix.

\paragraph{\textbf{Width}} 
In continuous sign language videos, individual signs span over multiple frames; hence, for analyzing the representation of each sign, we exploit its spatiotemporal properties by modeling it with a multi-scale STGCN. We use the kernel window in the Temporal Gated Convolution layer of the STGCN to obtain these segments. We consider a given video of length $N$, channels or dimension size $c$, and the number of 3d joints $v$. We define a 1D convolution kernel $\tau$ which maps the input chunks to a single output $Y$ \cite{yu2018spatio}.

\subsection{Training and Network Details}
Our proposed network is trained using the Pytorch framework \cite{pytorch} and follows the FAIRSEQ transformer setup \cite{ott-etal-2019-fairseq}.
We use Adam optimizer \cite{KingmaB14} with a batch size of 32 and initialize the learning rate to $10^{-4}$ with a weight decay of $10^{-4}$. The number of decoder attention heads was set to 12 and the best head was selected to analyze the output attention weights of each modality. The model was trained on a single NVIDIA RTX 3090 Ti processor.
\subsection{Evaluation Metrics}
We use the two popular language evaluation metrics to access our performance: BLEU (BiLingual Evaluation Understudy) \cite{10.3115/1073083.1073135} and ROUGE-L \cite{lin-och-2004-automatic}. BLEU computes a modified n-gram precision where for each candidate n-gram a maximum corresponding reference is counted. ROUGE-L was usd to measure the sentence-wise similarity based on the longest common sub-sequence statistics between a candidate translation and a set of reference translations.


\section{Experiments and Results}
\label{exp}
The proposed model is evaluated on the Phoenix2014T and ASLing datasets, and both quantitative and qualitative (explainable) results are reported. We experiment with different settings of input modalities on the model, reporting results on manual markers only, non-manual markers only, and manual + non-manual markers. The attention weights for each modality is dissected and our findings are presented via the extracted attention plots

\subsection{Quantitative}

We test the proposed model with benchmark Phoenix2014T dataset and achieve a BLEU4 \cite{papineni-etal-2002-bleu} score of $11.27$ on the task of Sign Language translation; the complete results are shown in Table Table \ref{table:gsl}. Table \ref{table:kysymys} shows the performance scores on the ASLing dataset. Due to its noisy nature and smaller size, we fine-tune the model on ASLing by transfer learning from Phoenix2014T. Recall that the aim of this work is to understand the structure and roles of different modalities in Sign Language Understanding.

\begin{table}[H]
\centering
\begin{center}
\begin{tabular}{|P{2.0cm}|P{1cm}|P{1.7cm}|P{1.8cm}|P{1cm}|P{1cm}|P{1cm}|P{1cm}|P{1.1cm}|}
\hline
\textbf{Method}                                                                                       & \textbf{Width}         & \textbf{Modality}                                              & \textbf{Feature}                                                     & \textbf{Bleu-1} & \textbf{Bleu-2} & \textbf{Bleu-3} & \textbf{Bleu-4} & \textbf{Rouge-L} \\ \hline
Sign2Text \cite{8578910}                                                                                            & -                      & Manual                                                         & -                                                                    & 32.24           & 19.03           & 12.83           & 9.58            & 31.80            \\ \hline
MCT \cite{10.1007/978-3-030-66823-5_18}                                                                                                  & -                      & Multi-channel                                                  & -                                                                    & -               & -               & -               & 18.51           & 43.57            \\ \hline
Skeletor \cite{9522847}                                                                                             & -                      & Manual                                                            & \begin{tabular}[c]{@{}c@{}}2d to 3d \\ lifting\end{tabular}          & 31.86           & 19.11           & 13.49           & 10.35           & 31.80            \\ \hline
\multirow{4}{*}{\begin{tabular}[c]{@{}c@{}}TSPNet \cite{li2020tspnet}\\ (Single)\end{tabular}}                            & \{8\}                  & -                                                              & RGB \cite{8099985}                                                                  & 30.29           & 17.75           & 12.35           & 9.41            & 28.93            \\ \cline{2-9} 
                                                                                                      & \{9\}                  & -                                                              & RGB \cite{8099985}                                                                 & 23.87           & 15.49           & 11.08           & 8.71            & 24.7             \\ \cline{2-9} 
                                                                                                      & \{12\}                 & -                                                              & RGB \cite{8099985}                                                                 & 29.02           & 17.03           & 12.08           & 9.39            & 28.10            \\ \cline{2-9} 
                                                                                                      & \{16\}                 & -                                                              & RGB \cite{8099985}                                                                 & 35.52           & 20.33           & 14.75           & 11.61           & 32.36            \\ \hline
\multirow{5}{*}{\begin{tabular}[c]{@{}c@{}}Parallel \\ Cross-\\Attention\\ Decoder \\(Ours)\end{tabular}} & \multirow{5}{*}{\{9\}} & Manual                                                         & \begin{tabular}[c]{@{}c@{}}3d\\ keypoints\end{tabular}               & 29.40           & 17.23           & 11.98           & 9.82            & 23.4             \\ \cline{3-9} 
                                                                                                      &                        & Non-manual                                                     & \begin{tabular}[c]{@{}c@{}}3d\\ keypoints\end{tabular}               & 22.35           & 11.32           & 7.68            & 6.25            & 15.5             \\ \cline{3-9} 
                                                                                                      &                        & \begin{tabular}[c]{@{}c@{}}Manual + \\ Non-manual\end{tabular} & \begin{tabular}[c]{@{}c@{}}3d\\ keypoints\end{tabular}               & 29.81           & 17.15           & 11.65           & 9.87            & -                \\ \cline{3-9} 
                                                                                                      &                        & Manual                                                         & rotmat9d                                                             & 24.65           & 13.24           & 9.07            & 7.44            & 18.4             \\ \cline{3-9} 
                                                                                                      &                        & \begin{tabular}[c]{@{}c@{}}Manual + \\ Non-manual\end{tabular} & \begin{tabular}[c]{@{}c@{}}3d-keypoints \\ +\\ rotmat9d\end{tabular} & 32.79           & 19.91           & 13.7            & 11.27           & -                \\ \hline
\end{tabular}
\end{center}
\caption{Performance of proposed method on Phoenix2014T dataset}
\label{table:gsl}
\end{table}

The intuition behind using the Phoenix2014T dataset is to verify the model's end performance and guarantee that the attention weights are learned correctly. We experimented with different temporal width settings with STGCN, and based on the accuracy achieved, we selected the width to be used in the model. We also experimented with different feature extraction methods, i.e., 3d-keypoints and 3d rotation matrix (rotmat9d). The reported numbers for non-manual markers for the ASLing dataset were substandard due to its noisy nature. Although there were several images with clear, full-frontal face views, the number of face images with poor lighting or pose conditions dominated. Nevertheless, we were still able to identify patterns consistent with the Sign Language semantics. We believe that the frames' quality can directly impact the feature quality; hence, the model attends poorly when the image quality is low, as shown in Figure \ref{fig:face-crops}. 

\begin{table}[H]
\centering
\begin{center}
\begin{tabular}{|l|l|l|l|l|l|l|}
\hline
\textbf{Method}                                                                       & \textbf{Modality}                                             & \textbf{Feature}                                                                & \textbf{Bleu-1} & \textbf{Bleu-2} & \textbf{Bleu-3} & \textbf{Bleu-4} \\ \hline
CFDF                                                                                  & Manual                                                        & \begin{tabular}[c]{@{}l@{}}2d-keypoint +\\ Optical FLow +\\ ResNet\end{tabular} & 22.39           & 15.96           & 13.56           & 12.25           \\ \hline
\begin{tabular}[c]{@{}l@{}}Parallel \\ Cross-Attention \\ Decoder (Ours)\end{tabular} & \begin{tabular}[c]{@{}l@{}}Manual +\\ Non-manual\end{tabular} & \begin{tabular}[c]{@{}l@{}} rotmat9d\end{tabular}               & 21.82            & 16.08            & 13.67            & 12.30            \\ \hline
\end{tabular}
\end{center}
\caption{Performance of proposed method on ASLing Transfer \\ learned on Phoenix2014T}
\label{table:kysymys}
\end{table}

\subsection{Qualitative}

\begin{figure}[H]

  \centering
  \includegraphics[width=0.85\linewidth]{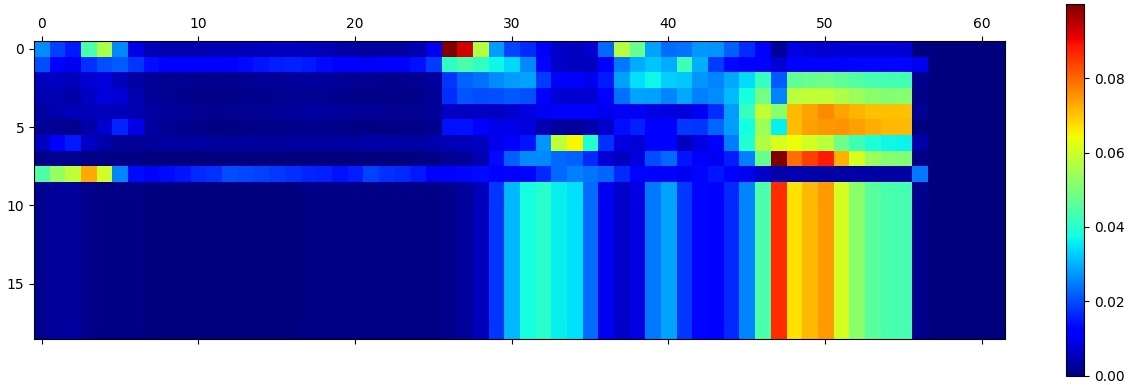}
  \includegraphics[width=0.85\linewidth]{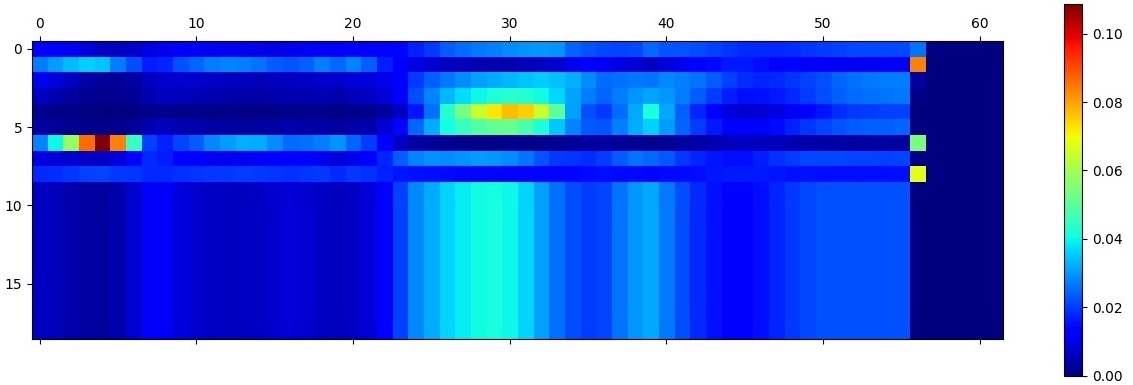}
  \caption{Learned attention weights for Phoenix2014T dataset. Top: attention weights based on the input manual markers (body features); bottom: attention weights based on the input non-manual markers (facial features)}
  \label{fig:gsl-atn-body}
\end{figure}

We observe the behavior of our proposed model by plotting the learned attention weights for each modality. The correlation is strong between a sequence frame and the decoder output token when that frame consists of strong facial feature movements that influence the output task. This indicates that along with body motions, often when signing, facial features can also contribute significantly to the output task (SLT in this case). 

\begin{figure}[]
  \centering
  \includegraphics[width=0.85\linewidth]{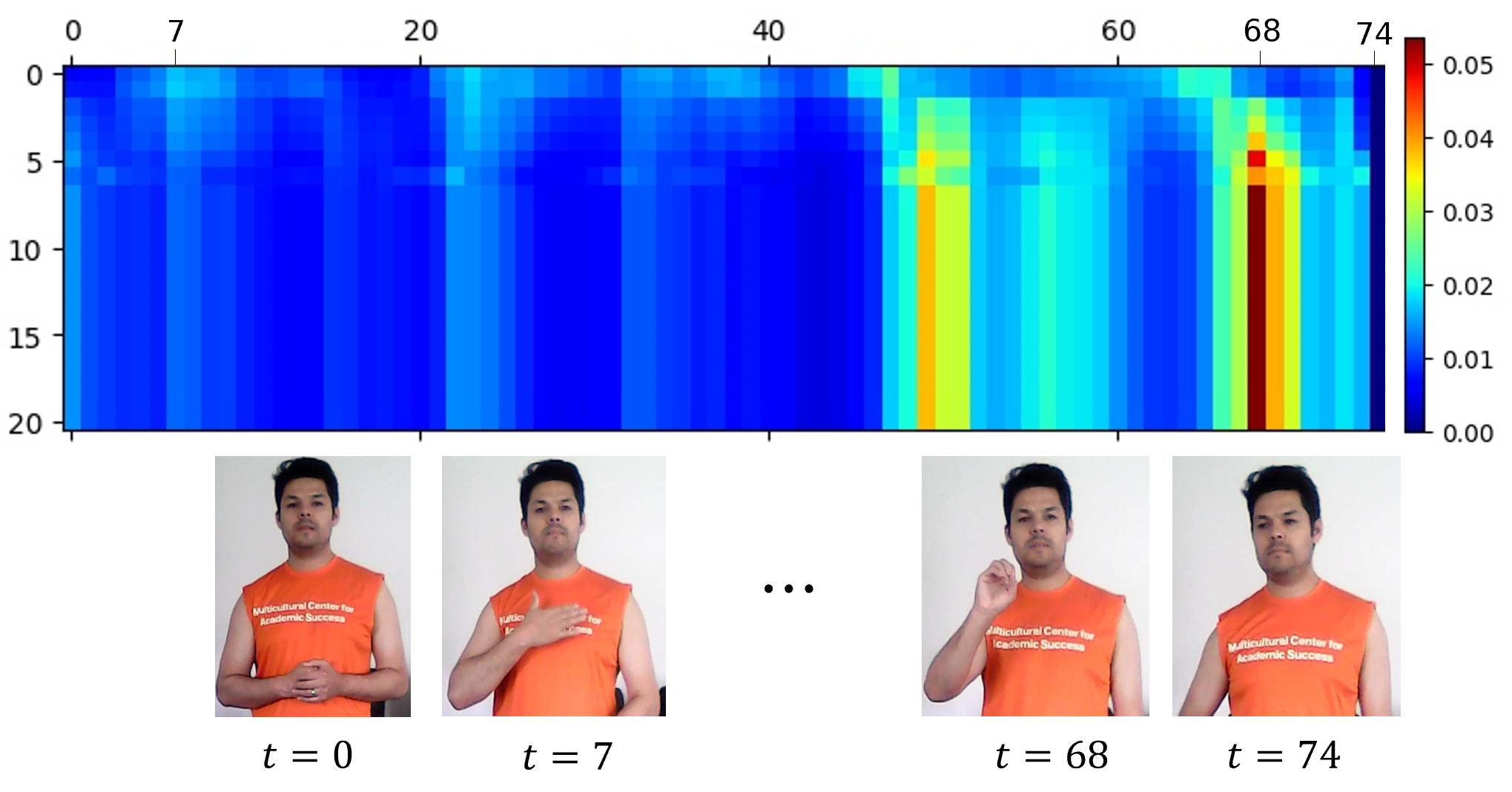}
  \includegraphics[width=0.85\linewidth]{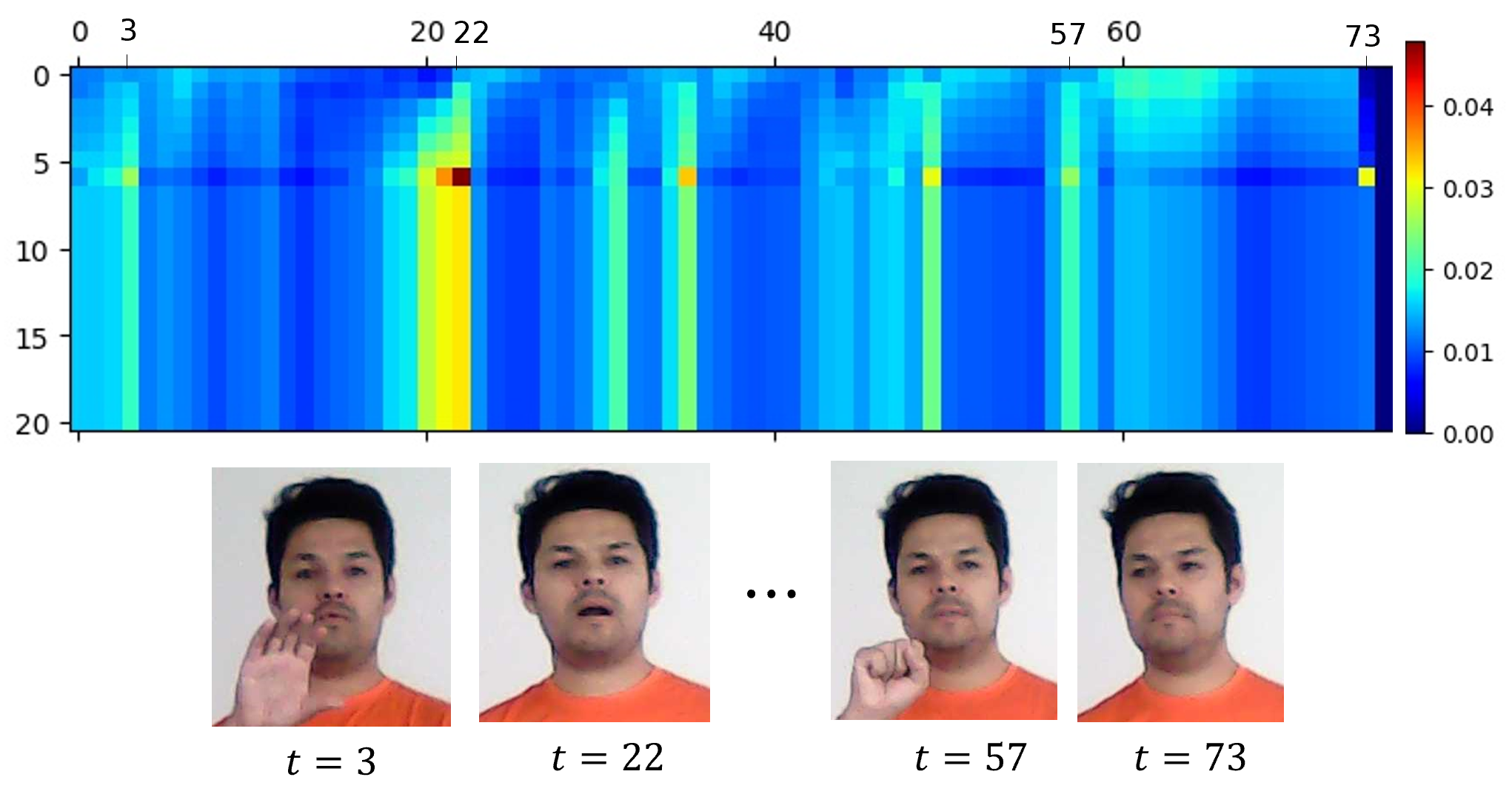}
  \caption{Learned attention weights for ASLing dataset. Top: attention weights based on the input manual markers (body features); bottom: attention weights based on the input non-manual markers (facial features).}
  \label{fig:asl-atn-body}
\end{figure}

Figure \ref{fig:gsl-atn-body} shows the attention weights for one sample from the testing set of the Phoenix2014T dataset. The x-axis of the plot represents the input sequence frames, and the y-axis represents the decoder output tokens. The darker blocks in Figure \ref{fig:gsl-atn-body} (bottom) demonstrate that the non-manual input features (facial expressions) are attending more in the correlations to the output tokens. 

\FloatBarrier\section{Discussion and Conclusion}

\paragraph{Fusion Techniques} Most of the previous works using multi-modalities have used early or late fusion of features. In the early fusion method, the features from different modalities are fused to create a single representation of the sequence. This implicitly helps the translation model build meaningful contextual relationships between the decoder output token and fused features. Although this model works well for fusing multimodal inputs, it cannot determine the extent of influence of each of the input modalities. 
Camgoz et al.\cite{10.1007/978-3-030-66823-5_18} noted that performing this type of late fusion does not always yield better results. Contrary to this, our proposed model performs fusion at decoding time, not to learn about feature representations but to extract learned attention weights for each modality.

We also show qualitative results on ASLing dataset in Figure \ref{fig:asl-atn-body}. The plots show the decoder output tokens plotted against the sequence frames. These plots show the attention weights for the phrase: \textbf{\textit{"Christmas is my favorite season!"}}. For the duration of the input sequence in this example, based on the attention weights, we see that the facial features attend to the output token a little more than the body features. This behavior can be seen when the signer asks a question or conveys excitement or enthusiasm. A larger attention weight value for a particular modality at time $t$ indicates that this feature contributed more to constructing the context representation between encoder and decoder outputs at time $t$.

\paragraph{Performance Metrics} Recent works have achieved better results than the proposed method on Phoenix-2014T, have used gloss annotations to supervise their training \cite{guan2024multistreamkeypointattentionnetwork}, \cite{chen2022two}.  Needless to say, it is assumed that the multi-modality methods are expected to give a better performance because of the added modality. However, the performance of adding a modality can not always be guaranteed. Many factors contribute to this; in the case of ASLing, the dataset was collected in the wild (real-life setting) and is noisy Figure  \ref{fig:face-crops} (Bottom). On the contrary, the Phoenix2014T dataset was collected in a more constrained environment where the participants wore dark clothing to contrast with the background; the environment also had controlled lighting - See Figure \ref{fig:face-crops} (Top). This can pose a challenge when analyzing the ASLing dataset. Additionally, non-manual markers convey more than just facial expressions they also construct the grammatical meaning of a sign. This can be a complex structure to decode and understand if the rules of a language are not learned.

\FloatBarrier\subsection*{Conclusion}

In this paper, in an attempt to develop an influence model in a multimodal architecture, we introduced a dual encoder model, along with a parallel cross-attention decoder, to study the contributions of manual and non-manual features in Sign Language translation. Through the parallel cross-attention mechanism, we are able to retrieve the attention weights for individual modalities, and while the underlying goal is Sign Language translation (proven via quantitative measures), the proposed parallel cross-attention mechanism proved exceptionally useful in estimating the contribution of influence that each modality had on the decoder output during inference. This allowed us to measure the influence of facial expressions on sign translation for different types of signed input phrases. This attribute of the model is its major distinguishing factor among other existing Transformer-based architectures.

%
%
%
\bibliographystyle{splncs04}
\bibliography{mybibliography}

\end{document}